\documentclass[11pt,a4paper]{article}
\usepackage[hyperref]{acl2021}
\usepackage{times}
\usepackage{latexsym}
\usepackage{amsmath}
\usepackage{mathrsfs}
\usepackage{graphicx}
\usepackage{graphics}
\usepackage{url}
\usepackage{xr}
\usepackage{booktabs}
\usepackage{balance}
\usepackage{caption}
\usepackage{subcaption}
\usepackage{enumitem}
\usepackage{pgfplots}
\usepackage{tikz}
\usepackage{xcolor}
\usepackage{xspace}
\usepackage{makecell}

\definecolor{g-red}{HTML}{DB4437}
\definecolor{g-blue}{HTML}{4285F4}
\definecolor{g-green}{HTML}{0F9D58}
\definecolor{g-yellow}{HTML}{F4B400}
\definecolor{g-orange}{HTML}{FF9800}
\definecolor{g-grey}{HTML}{9E9E9E}
\definecolor{g-black}{HTML}{000000}

\usepackage[T1]{fontenc}
\usepackage[utf8]{inputenc}
\usepackage{microtype}
\aclfinalcopy
\usepackage{microtype}

\newcommand{\bert}[1]{\textsc{BERT}$_{#1}$\xspace}
\newcommand{\nfr}{$\mathcal{R}_{NF}$\xspace}

\title{Regression Bugs Are In Your Model! \\ Measuring, Reducing and Analyzing Regressions In NLP Model Updates}

\author{Yuqing Xie,$^{1,2}$ {\bf Yi-an Lai},$^1$
{\bf Yuanjun Xiong,}$^1$ {\bf Yi Zhang,}$^1$ \and {\bf Stefano Soatto}$^1$\vspace{0.1cm}\\
$^1$ Amazon AWS AI,
$^2$ University of Waterloo
}
\date{}

\begin{document}
\maketitle

\begin{abstract}
Behavior of deep neural networks can be inconsistent between different versions.
Regressions\footnotemark during model update are a common cause of concern that often over-weigh the benefits in accuracy or efficiency gain. 
This work focuses on quantifying, reducing and analyzing regression errors in the NLP model updates.
\footnotetext{Here \textit{regression} refers to bugs in software testing instead of the statistical estimation method.}
Using \emph{negative flip rate} as regression measure, we show that regression has a prevalent presence across tasks in the GLUE benchmark.
We formulate the regression-free model updates into a constrained optimization problem, and further reduce it into a relaxed form which can 
be approximately optimized through knowledge distillation training method. We empirically analyze how model ensemble reduces regression.
Finally, we conduct \textsc{CheckList} behavioral testing to understand the distribution of regressions across linguistic phenomena, and the efficacy of 
ensemble and distillation methods. 
\end{abstract}

\section{Introduction}

Regression-free model update is a desirable system property which guarantees interoperability of a new system with a legacy version, also known as backward compatibility. 
Regression occurs when the newly updated system stops functioning as intended. 

As advances in deep learning spark industrial applications in AI areas such as natural language processing, the long-term maintenance of such systems is becoming ever more challenging. While models with complex neural architectures and huge parameter space continue to reach higher accuracy, the lack of interpretability and functional decomposibility in these models make it infeasible to apply traditional software regression testing methods such as unit tests. As result, validating and mitigating regressions during model update is often a long and painful engineering process, which often over-shadows the benefits of a new model. 

The model regression issue in deep learning first comes into sight in \citet{Shen2020TowardsBR}, where they inspect compatible representation learning for image retrieval.
\citet{yan2020positive} proposed the positive-congruent training (PCT) for image classification that minimizes prediction errors and model regression at the same time.
To our best knowledge, the model update regression has not been studied on NLP tasks.

\begin{figure}[t!]
\centering
\includegraphics[width=0.48\textwidth]{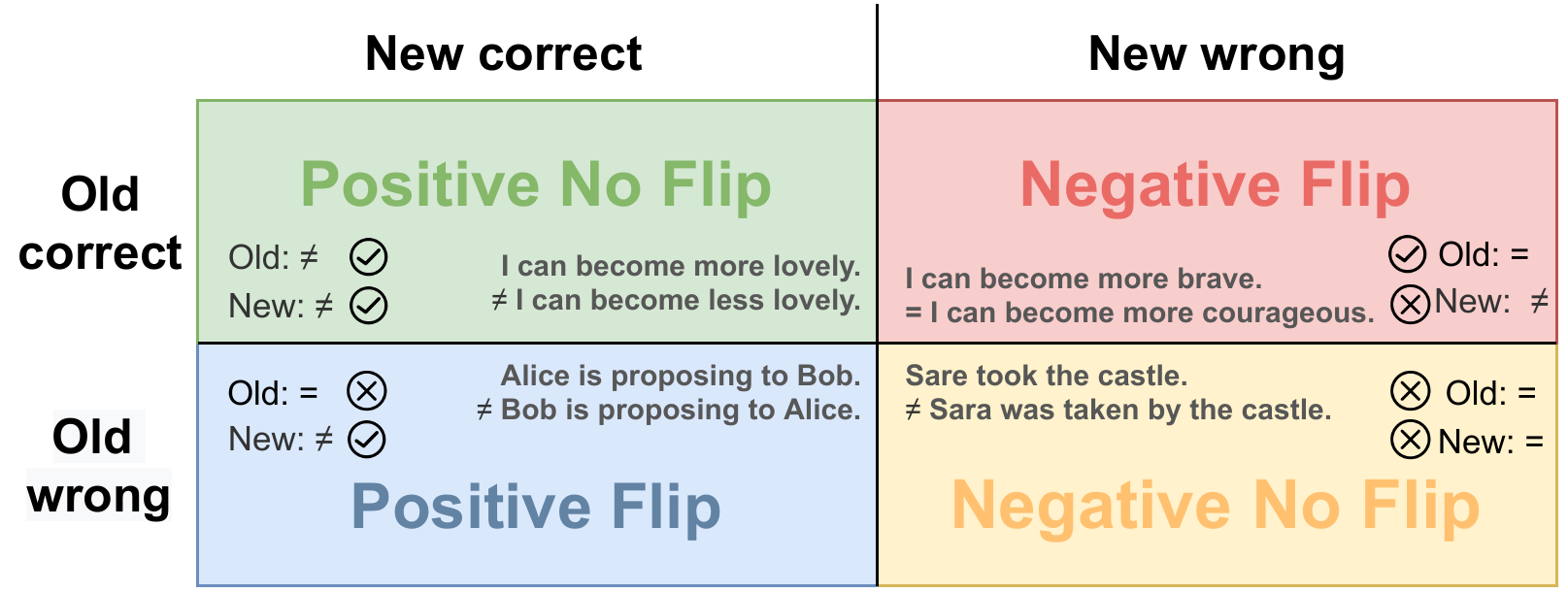}
\caption{Prediction flip scenarios on NLP classification tasks when updating from old to new models. Examples from paraphrase classification task.}
\label{fig:regression}
\end{figure}

Following \citet{yan2020positive}, in this work we measure the model update regression in NLP by \emph{negative flips}.
In Figure \ref{fig:regression}, we demonstrate prediction flip scenarios. Negative flips are shown in the upper-right quadrant where the old model makes correct predictions and the new model predictions are wrong. 
As we will show in Section~\ref{sec:measure}, regression are prevalent in NLU model updates even with the slightest changes in the new model training process. 

To develop a model with minimum regression, we first formulate the learning task into a constrained optimization problem by taking the regression-free conditions as constraints.
We apply the Lagrangian relaxation to bring the regression-free constraint into the optimization objective as an additional penalty loss, and provide approximate solution via knowledge distillation.
\citet{yan2020positive} also observed that model ensemble can also reduce negative flips without explicit input from the old model. We evaluate both distillation and ensemble based methods on a diverse set of NLP tasks. 

To further understand how the above methods contribute to reducing it, we utilize \textsc{CheckList} \citep{ribeiro-etal-2020-beyond} to quantify linguistic behavioral changes before and after applying proposed methods.
We find that regressions are prevalent in NLP tasks, and their distribution correlates with different linguistic phenomena. 

Our main contributions are as follows:
\begin{itemize}[itemsep=4pt,topsep=4pt,parsep=0pt,partopsep=0pt]
    \item We provide empirical evidence to show that the model update regression occurs across text classification tasks in NLP;
    \item We formulate the regression-free model updates into a constrained optimization problem, and further reduce into a relaxed form which can be approximately optimized through knowledge distillation training method;
    \item We also explore the model ensemble as another method to reduce regression, and analyzed its efficacy;
    \item We analyze the source of the regressions in NLP tasks through linguistic behavioural testing, compare reduction in both distillation and ensemble methods.
    
\end{itemize}

\begin{table*}[t!]
\centering
\resizebox{0.9\textwidth}{!}{
\begin{tabular}{lc|rrrr}
\toprule
 && \textbf{CoLA}  & \textbf{MRPC} & \textbf{QQP} & \textbf{MNLI-m}  \\
\midrule
\multicolumn{1}{c}{\textbf{Train size}}&  & 8.6k   & 3.7k  & 360k  & 390k  \\
\multicolumn{1}{c}{\textbf{Dev size}}& & 1k   & 0.4k  & 40k   & 9.8k   \\
\midrule
\textbf{Old: \bert{base}}& Acc &82.84\%&	86.03\%&	90.76\%&	83.82\%	\\
\midrule
\textbf{$\rightarrow$ \bert{base}}& Acc & 83.80\%(+0.96\%)&	86.03\%(-0.00\%)&	90.56\%(-0.20\%)&	83.55\%(-0.27\%)\\
& ${\mathcal{R}_{NF}}$& $\hookleftarrow$3.36\%& $\hookleftarrow$	    4.17\% &	 $\hookleftarrow$2.32\%& $\hookleftarrow$	3.56\%  \\
\hline
\textbf{$\rightarrow$ \bert{large}}& Acc &85.43\%(+2.59\%)&	87.75\%(+1.72\%)&	91.11\%(+0.35\%)&	86.10\%(+1.97\%)\\
& ${\mathcal{R}_{NF}}$  & $\hookleftarrow$ 3.16\%        & $\hookleftarrow$ 5.88\%    &  $\hookleftarrow$2.82\%    & $\hookleftarrow$ 3.95\%\\
\hline
\textbf{$\rightarrow$ \textsc{RoBERTa}$_{base}$}& Acc & 84.85\%(+2.01\%)&	89.22\%(+3.19\%)&	91.25\%(+0.52\%)&	87.58\% (+4.42\%) \\
& ${\mathcal{R}_{NF}}$  & $\hookleftarrow$4.67\%& $\hookleftarrow$	4.66\%&	 $\hookleftarrow$1.98\%& $\hookleftarrow$	2.64\%\\
\hline
\textbf{$\rightarrow$ \textsc{ELECTRA}$_{base}$} &Acc &85.81\%(+2.97\%)&	86.03\%(+0.00\%)&	91.35\%(+0.59\%)&	88.87\%(5.05\%)\\
&${\mathcal{R}_{NF}}$   & $\hookleftarrow$ 5.18\%   & $\hookleftarrow$ 5.39\%  & $\hookleftarrow$ 3.20\%  &  $\hookleftarrow$3.50\%\\
\hline
\textbf{$\rightarrow$\textsc{ALBERT}$_{base}$} &Acc &76.51\%(-6.33\%)&	86.27\%(+0.24\%)&	90.73\%(-0.03\%)&	85.26\%(+1.44\%)\\
&${\mathcal{R}_{NF}}$& $\hookleftarrow$10.74\%&$\hookleftarrow$	6.86\%&$\hookleftarrow$	3.78\%&$\hookleftarrow$	5.22\%\\
\bottomrule
\end{tabular}
}
\caption{Regression measurement when performing different model updates on GLUE benchmark. Old: \bert{base} is our base version of model and we update it to other new models. We also show the accuracy gain in the parenthesis. All numbers are $5$-seed average.\footnotemark
}
\label{table:regressions}
\end{table*}

\section{Measuring Regression in NLP Model Update}
\label{sec:measure}

In this section, we first formulate the measure of model update regression on classification tasks.
Then we benchmark on GLUE tasks \citep{wang2018glue} and show that there is a prevalent presence of regression when updating models in NLP.

\subsection{Regression Measurement on Classification Tasks}

Similar to software regression testing, we need to collect a group of test cases when measuring regression.
We start from a regression set:
$\mathcal{D}_{reg}=\{x_i, y_i\}_{i=1}^{N}, \: y_i \in \{l_1, l_2, ..., l_C\}$, where $l_i$ is the $i$-th label and $C$ is the number of classes.
In practice, we can use the development set or compile a collection of critical use cases as $\mathcal{D}_{reg}$.

In a classification task, given a input $x_i$, a neural network model $f$, parameterized by $\phi$, approximates the posterior probabilistic distribution $p(y_i | x_i)$ over all possible labels:
$\vec{f}_{\phi}(x_i) = (p_{\phi}(y=l_1|x_i), ..., p_{\phi}(y=l_C|x_i))^{\top}$.
To simplify, we denote the final prediction of a model to be $f_{\phi}(x) = arg \max_{l_j} p_{\phi}(l_j|x) $.

The regression $\mathcal{R}_{NF}$ between two models $f_{\phi_{old}}$ and $f_{\phi_{new}}$ on $\mathcal{D}_{reg}$ can be defined as the portion of negative flip cases:
\begingroup\makeatletter\def\f@size{10}\check@mathfonts
\begin{equation}
\mathcal{R}_{NF}(\mathcal{D}_{reg}, \vec{f}_{\phi_{old}}, \vec{f}_{\phi_{new}})
=\frac{|\{x|f_{\phi_{old}}=y, f_{\phi_{new}}\neq y\}|}
{|\mathcal{D}_{reg}|}.
\end{equation}
\endgroup
We use negative flip $\mathcal{R}_{NF}$ as our regression measurement for classification tasks.
Lower $\mathcal{R}_{NF}$ for a new models means better compatibility with the old model.

\subsection{Benchmark Severity of Regression}

The success of Transformer \citep{NIPS2017_3f5ee243} and BERT \citep{devlin-etal-2019-bert} have made pre-training then fine-tuning a standard paradigm in NLP systems.
When updating these systems, differences can come from various aspects:
\begin{itemize}[itemsep=3pt,topsep=3pt,parsep=0pt,partopsep=0pt]
    \item Changes in the fine-tuning hyperparameters (e.g. random seed, learning rate schedule, epoch, etc.)
    \item Changes in model size or architecture (e.g. from \bert{base} to \bert{large})
    \item Changes in pre-training procedure or objective (e.g. \textsc{BERT} to \textsc{RoBERTa} \citep{Liu2019RoBERTaAR}, to $\textsc{BERT}_{\text{whole-word-masking}}$ or to $\textsc{ELECTRA}$ \citep{Clark2020ELECTRA:})
    \item Changes in pre-trained model architecture (e.g. $\textsc{BERT}$ to $\textsc{ALBERT}$ \citep{Lan2020ALBERTAL:})
\end{itemize}

\noindent While accuracy or efficient improvements are strong motivations for these model updates, they could also introduce behavioral incongruence when compared to the previous model.
To benchmark the severity of regression, we apply a general setup: Fine-tune various pre-trained language models (LM) on GLUE and calculate $\mathcal{R}_{NF}$ when updating from \bert{base} to other LMs .
We use dev sets as $\mathcal{D}_{reg}$.
Results in Table \ref{table:regressions} show that:
\begin{enumerate}[itemsep=3pt,topsep=3pt,parsep=0pt,partopsep=0pt]
    \item \emph{Model update regression is prevalent on NLU tasks.} A minimum of $1.98\%$ ${\mathcal{R}_{NF}}$ is observed across diverse classification tasks and model update scenarios, while the average accuracy gain is only $1.4\%$.
    \item \emph{Minor changes such as random seeds can introduce significant regression.} 
    Shown in $\rightarrow$\bert{base}, even when we only alter the initialization random seed, this can lead to up to $3.56\%$ negative flip.
    \item \emph{Negative flip rates are often much higher than the accuracy gains.}
    When updating to \bert{large} on QQP, $\mathcal{R}_{NF}$ is about $8\textsc{x}$ the accuracy gain. 
    This implies reducing error rate alone does not ensure the decrease in regression.
    \item \emph{Pre-training objective or architecture updates often lead to higher regressions than those caused by model size or random seeds.} 
    The regressions are higher when updating to $\text{ALBERTA}$, compared with updating to a larger model \bert{large} or a different random seed.
    This implies systematic regression could be introduced if the backbone models are different.
\end{enumerate}

\section{Reducing Regression in Model Update}\label{sec:method}
\footnotetext{Full results on GLUE can be found in Appendix A}

In this section, we first formulate regression-free model update as a constrained optimization problem, then further reducing it to a joint optimization objective 
combining the training loss on the original task and a distillation loss with respect to the old model's behavior.

Unlike typical optimizations in neural model training where we minimizes a loss function on a training set, the regression-free model update requires the model to learn the target task as well as comply with conditions posed by the old model.
We can cast the regression-free model update as a constrained optimization problem by writing down the classification loss as the optimization objective and the regression-free conditions as constraints:
\begin{equation}\label{eq:origin}
\begin{split}
    \min_{\phi_{new}} \  &\sum_{x\in \mathcal{D}_{train}}\mathcal{L}_{CE}(x, \phi_{new})\\
    \textit{s.t.} \  & \mathcal{R}_{NF}(f_{\phi_{old}}, f_{\phi_{new}}, \mathcal{D}_{reg})=0.
\end{split}
\end{equation}
\noindent where $\mathcal{D}_{train}, \mathcal{D}_{reg}$ represent the training and regression sets, respectively.

The constraint in Equation \ref{eq:origin} asks for zero regression on $\mathcal{D}_{reg}$.
It would be difficult to ensure the constraint is satisfied along the model training.
We instead relax the hard constraint into a soft inequality condition that allows the regression measure to be less than a constant $\mathcal{C}$:
\begin{equation}
\begin{split}
    \min_{\phi_{new}} \  &\sum_{x\in \mathcal{D}_{train}}\mathcal{L}_{CE}(x, \phi_{new})\\
    \textit{s.t.} \  & \mathcal{C} - \mathcal{R}_{NF}(f_{\phi_{old}}, f_{\phi_{new}}, \mathcal{D}_{reg}) \geq 0.
\end{split}
\end{equation}

Training a model directly with the regression-free constraint still remains difficult in that signals from old predictions are sparse and $\mathcal{R}_{NF}$ is non-differentiable.
Here, we propose two proxies of $\mathcal{R}_{NF}$ to measure regression in continuous space.

\vspace{4pt}
\noindent \textbf{Proxy from Prediction Probabilities.}
We use the KL divergence between the predicted probabilities of both models as one soft regression measure:
\begin{equation}
\begin{split}
&\mathcal{R}_{\text{KL-div}}(f_{\phi_{old}}, f_{\phi_{new}}, \mathcal{D}_{reg}) \\
=& \sum_{x\in\mathcal{D}_{reg}} D_{KL}(p_{\phi_{old}}(y|x)||p_{\phi_{new}}(y|x)).
\end{split}
\end{equation}

\vspace{2pt}
\noindent \textbf{Proxy from Deep Representations.}
We can also use the $l_2$ distance between models' sentence representations, e.g. \texttt{[CLS]} embedding in $\text{BERT}$ as another soft regression measure:
\begin{equation}
\begin{split}
&\mathcal{R}_{l_2}(f_{\phi_{old}}, f_{\phi_{new}}, \mathcal{D}_{reg}) \\
=& \sum_{x\in\mathcal{D}_{reg}} l_2(\vec{f}_{\phi_{old}}(x), \vec{f}_{\phi_{new}}(x)).
\end{split}
\end{equation}
A linear projection is used to align the representations if they initially lie in different spaces.

\vspace{4pt}
\noindent \textbf{Reduce to Knowledge Distillation.}
Finally, we apply the Lagrangian relaxation to bring the regression-free constraint into the optimization objective as an additional penalty loss:
\begin{equation}
\begin{split}
\min_{\phi_{new}} \  &\sum_{x\in \mathcal{D}_{train}}\mathcal{L}_{CE}(x, \phi_{new})\\  
    -& \ \alpha * (\mathcal{C} - \mathcal{R}_{soft}(f_{\phi_{old}}, f_{\phi_{new}}, \mathcal{D}_{reg})),
\end{split}
\end{equation}

\noindent where $\alpha$ is a positive penalty scaling parameter and $\mathcal{R}_{soft}$ can be chosen from $\mathcal{R}_{\text{KL-div}}$ or $\mathcal{R}_{l_2}$.
Then, the above optimization problem can be cast into a joint learning of the original target task and knowledge distillation from the old model.
The distillation loss acts as a surrogate of the model update regression measure.
The joint learning process minimizes this term as an approximation of minimizing the overall model update regression.

\begin{table*}
\footnotesize
\centering
\resizebox{0.99\textwidth}{!}{
\begingroup
\renewcommand{\arraystretch}{1.15}
\begin{tabular}{cc|ccccccc|c}
\toprule
 && \textbf{CoLA} & \textbf{SST-2} & \textbf{MRPC} & \textbf{QQP} & \textbf{MNLI-m}  & \textbf{QNLI} & \textbf{RTE} &\textbf{Average}\\
\midrule
\textbf{Train size}&  & 8.6k & 67k   & 3.7k  & 360k  & 390k   & 100k  & 2.5k \\
\textbf{Dev size}& & 1k  & 0.9k  & 0.4k  & 40k   & 9.8k   & 5.5k  & 0.3k \\
\midrule
\textbf{Old: \bert{base}} & Acc 
&82.26\%    &	91.17\% & 86.03\%   &	90.76\% &	83.82\% &91.07\%    &67.15\%&84.61\%\\
\midrule
\textbf{$\rightarrow$\bert{base}} - Baseline & Acc 
& 82.93\%   &	91.63\% & 86.03\%   &	90.56\% &	83.55\% &90.65\%    &63.18\%&84.08\%\\
& ${\mathcal{R}_{NF}}$
& 4.41\%    &	1.95\%  &	4.17\%  &	2.32\%  &	3.56\%  &	2.35\%  &	11.43\%     &4.31\% \\
\textbf{$\rightarrow$\bert{base}} - Distillation & Acc 
& \textbf{84.47}\%  & \textbf{92.09}\%  &\textbf{87.01}\%   &\textbf{91.14}\%  &83.77\%  &91.16\%  &68.95\%&\textbf{85.81}\%\\
& ${\mathcal{R}_{NF}}$
& \textbf{1.92}\%    &\textbf{0.80}\%  &1.72\%  &	1.69\%  &4.32\% &2.47\%&8.30\%    &2.99\%  \\
\textbf{$\rightarrow$\bert{base}} - Ensemble & Acc 
& 82.17\%  & 91.63\%  &86.03\%   &91.06\%  &\textbf{84.35}\%  &\textbf{91.62}\%  &\textbf{70.76}\%&85.37\% \\
 & ${\mathcal{R}_{NF}}$
& 2.59\%    &0.92\%  &\textbf{1.23}\%  &	\textbf{1.18}\%  &\textbf{1.66}\% &\textbf{1.06}\%&\textbf{4.69}\%   &\textbf{1.90}\%  \\
\midrule
\textbf{$\rightarrow$\bert{large}} - Baseline& Acc 
&85.62\%    &	92.89\% &	87.75\% &	91.11\% &	86.10\% &		92.53\% &	66.43\%&86.06\%\\
& ${\mathcal{R}_{NF}}$        
& 2.68\%    & 1.72\%    & 5.88\%    & 2.82\%    & 3.95\%       & 2.64\%    & 12.27\%    &4.57\%  \\
\textbf{$\rightarrow$\bert{large}} - Distillation & Acc 
&\textbf{85.62}\%    &92.89	\%  &88.73\%    &91.50\% &86.73	\%   &92.15\%  &	 \textbf{73.65}\%&\textbf{87.33}\%\\
& ${\mathcal{R}_{NF}}$   
&\textbf{2.49}\% &\textbf{1.26}\%    &2.45\%    & 2.46\%    &3.76\% &2.54 \%    & \textbf{5.42}\%   &\textbf{2.91}\%  \\
\textbf{$\rightarrow$\bert{large}} - Ensemble & Acc 
& 84.95\%  & \textbf{93.12}\%  &\textbf{89.46}\%   &\textbf{91.66}\%  &\textbf{87.05}\%  &\textbf{93.08}\%  &67.87\%&86.74\%\\
& ${\mathcal{R}_{NF}}$
& 2.78\%    &1.61\%  &2.45\%  &	\textbf{2.20}\%  &\textbf{3.24}\% &\textbf{2.27}\%&10.83\%&3.63\%\\
\bottomrule
\end{tabular}
\endgroup
}
\caption{Results of fine-tuning with distillation and ensemble on GLUE benchmark. Baseline denotes directly fine-tuning new pre-trained models on target tasks. We show the distillation results with $\mathcal{R}_{\text{KL-div}}$, and the ensemble results with 5 model majority vote. Due to page limitation, we only show the matched results on MNLI \citep{williams-etal-2018-broad}.}
\label{table:CVP}
\end{table*}

\section{Experiments}\label{sec:exp}

\subsection{Implementation Details}

Since we usually update models from elementary ones to improved ones, in the experiments we take origin \bert{base}(12-layer, 768-hidden, 12-heads, 110M parameters) \citep{devlin-etal-2019-bert} as the old model's backbone and update it to a homogeneous model, e.g. \bert{base} with different fine-tuning random seeds or parameters, or a heterogeneous models with improvements such as \bert{large}(24-layer, 1024-hidden, 16-heads, 340M parameters). 
We fine-tune the pre-trained LMs without any constraint as our baselines. 
We use the GLUE datasets to benchmark the effectiveness of proposed techniques.
Pre-trained model artifacts and the GLUE dataset processing procedures are brought from Hugging Face\footnote{\url{https://huggingface.co}} and experiments are done in PyTorch \cite{NEURIPS2019_9015} with Tesla V100 GPUs.
Cross-entropy is used for fine-tuning on target tasks with batch size $16$ for $4$ to $6$ epochs. 
The learning rate is searched among $2\text{e}^{-5}$, $3\text{e}^{-5}$ and $5\text{e}^{-5}$. 

During joint training of classification and knowledge distillation, we take the fine-tuned old models as the teacher, and distill with batch size $16$ for $6$ to $8$ epochs. 
We set $\mathcal{D}_{reg}=\mathcal{D}_{train}$ when training models with the constraint and use $\mathcal{D}_{reg}=\mathcal{D}_{dev}$ for reporting results.
To encourage constraint satisfaction and reduce regression, we only include the distillation penalty into our loss on the examples where the current model makes negative flips.

\subsection{Ensemble}

\citet{yan2020positive} reported an intriguing finding on image classification tasks that model ensemble can reduce model update regressions without explicit regularization from the old model. This was attributed to the reduction of variance in ensemble model predictions, making it less prone to over-fitting and indirectly reducing regressions. Here we include model ensemble as an alternative approach to reduce regression, with further analysis on how ensemble reduces regression in Section \ref{sec:ensemble-analysis}.

\subsection{Main Results}

Table \ref{table:CVP} shows the efficacy of distillation method and model ensemble on reducing NLP classification task model update regressions. 
On average, the distillation method reduces $\mathcal{R}_{NF}$ by $30.6\%$ and $36.3\%$ while the ensemble method reduces $\mathcal{R}_{NF}$ by $55.9\%$ and $20.6\%$ when updating to \bert{base} and to \bert{large}, respectively.
Both distillation and ensemble methods can significantly bring down negative flips across GLUE tasks compared with the baselines.
The ensemble seems to work better when the old and new models share the same underlying pre-trained LM. In the update \bert{base}$\rightarrow$\bert{base}, the ensemble method outperforms the distillation on reducing the regression. 
On the other hand, the distillation method seems to be more effective on reducing regression under the heterogeneous model update setting. 
In the update \bert{base}$\rightarrow$\bert{large}, distillation reduce more regression, with especially large reductions on small datasets such as CoLA and SST-2.
We hypothesize that it's because the ensemble focuses on reducing the variance in model predictions, while distillation enables the explicit alignment in either probability distribution or representation space between the old and the new model.
When the new model is very different from the old one, it can implicitly align new model's behavior with the old one.

\subsection{Variants in Distillation Objective}
As introduced in Section \ref{sec:method}, we can have several variants of distillation loss to be used to constrain new model training on the old model. 
We explore and benchmark the following variants on the MRPC task:

\begin{itemize}[itemsep=3pt,topsep=3pt,parsep=0pt,partopsep=0pt]
    \item \textbf{Distillation - $\mathcal{R}_{\text{KL-div}}$, Logits} calculates the distillation loss as the KL divergence between the two Bernoulli distributions set by the old and new model prediction probabilities;
    \item \textbf{Distillation - $\mathcal{R}_{l_2}$, [CLS]} uses the \texttt{[CLS]} token embedding from the final layer as sentence representations and calculates the distillation loss as the Euclidean distance between the two vectors;
    \item \textbf{Distillation - $\mathcal{R}_{l_2}$, All [CLS]} also calculates the Euclidean distance between the old and new sentence representation vectors, but with concatenated \texttt{[CLS]} token embeddings from all layers instead of the final layer.
\end{itemize}

Pre-trained models could have different layers.
For \bert{base}$\rightarrow$\bert{large} in the \textbf{All [CLS]} setup, we align representations from \bert{large}'s even layers with the corresponding \bert{base} layers, e.g. $14$-th layer in \bert{large} is aligned with $7$-th in \bert{base}.

\begin{table*}[htbp]
\footnotesize
\centering
\resizebox{0.65\textwidth}{!}{
\begin{tabular}{l|cc|cc}
\toprule
\textbf{Old: \bert{base}}    & \multicolumn{4}{c}{\textbf{Acc}: 86.03\%}\\
\midrule
\textbf{New:} & \multicolumn{2}{c|}{\textbf{$\rightarrow$\bert{base}}} &\multicolumn{2}{c}{\textbf{$\rightarrow$\bert{large}}}\\
& {\textbf{Acc}} &{\textbf{$\mathcal{R}_{NF}$}}  &{\textbf{Acc}} &{\textbf{$\mathcal{R}_{NF}$}} \\
\midrule
\textbf{Baseline}  &86.03\%    & 4.17\%     &87.75\% &5.88\% \\
\midrule
\textbf{Distillation - $\mathcal{R}_{\text{KL-div}}$, Logits}& \textbf{87.01}\%    & 1.72\% &88.73\%    &2.45\% \\
\textbf{Distillation - $\mathcal{R}_{l_2}$, [CLS]}  &85.54\%    & 3.19\%    &88.73\%    &\textbf{2.21}\%    \\
\textbf{Distillation - $\mathcal{R}_{l_2}$, All [CLS]}      &85.54\%   &2.45\%      &87.99\%  &4.90\% \\
\midrule
\textbf{Ensemble}   &86.0\% &\textbf{1.23}\% &\textbf{89.46}\%    &2.45\%    \\
\bottomrule
\end{tabular}
}
\caption{Accuracy and regression results on MRPC with \bert{base}$\rightarrow$\bert{base} and \bert{base}$\rightarrow$\bert{large} updates using variants of distillation and ensemble methods. Baseline is fine-tuning with different random seeds. Model accuracy and negative flip rates are averaged across 5 seeds.}
\label{table:main-mrpc}
\end{table*}

Table \ref{table:main-mrpc} shows the results.
In the homogeneous setup, the most effective variant is to align the prediction probabilities via $\mathcal{R}_{\text{KL-div}}$, where it achieves up to $58\%$ $\mathcal{R}_{NF}$ reduction, i.e. from $4.17\%$ to $1.72\%$.
For $\mathcal{R}_{l_2}$ setup, aligning at all layers can further reduce $\mathcal{R}_{NF}$ compared with only aligning at the final layer. 
This implies a deeper alignment can help the new model more effectively learn to behave similarly as the old one when fine-tuning the same architecture with a different random seed.

In the heterogeneous setup, \textbf{$\mathcal{R}_{l_2}$, [CLS]} works the best for \bert{large} that achieves $62\%$ $\mathcal{R}_{NF}$ reduction, with $\mathcal{R}_{\text{KL-div}}$ having a comparable performance.
Overall, $\mathcal{R}_{\text{KL-div}}$ produces consistent regression reductions across different setups, which we pick it as our default setting in the distillation method.

From Table \ref{table:main-mrpc}, we can also observe that the deeper alignment seems to hurt $\mathcal{R}_{NF}$ in the heterogeneous update setup.
The reason might be that differences between pre-trained models are too significant.
The distillation with simple all-layer alignment could mess up pre-trained representations rather than effectively encourage new models to learn where the old model performs well.

Another interesting finding is that the simple model ensemble is a competitive solution comparing to the distillation.
In the \bert{base}$\rightarrow$\bert{base} setup, the ensemble even outperforms all the other distillation variants.
This is indeed a bit counter-intuitive as the distillation explicitly encourages the new model to pick up old models' correct predictions while the ensemble does not involve the old model in the process.
We conduct deeper analysis trying to understand on which aspects that these methods work to reduce the regression in the next section.

\section{Analyzing Regression in Model Updates}

In this section, we first analyze the model ensemble and present our hypothesis on how it reduces regression.
Next, we conduct behavioral testing across diverse linguistic phenomena to see where the reduced and remaining regressions reside. 

\subsection{Analysis of Updating to Model Ensemble}\label{lab:center}
\label{sec:ensemble-analysis}

\begin{figure}[t]
\centering
\begin{tikzpicture}
	\begin{axis}[
	width=8cm,
	height=5cm,
	legend cell align=left,
	legend style={at={(0.3,1.1)}, anchor=south west, font=\tiny},
	mark options={mark size=3},
	font=\scriptsize,
	xmin=0, xmax=8,
	ymin=0, ymax=60,
	xtick={1.0, 2.0, 3.0, 4.0, 5.0, 6.0, 7.0, 8.0},
	ytick={0, 10, 20, 30, 40, 50, 60},
	xmajorgrids=true,
	ymajorgrids=true,
	xlabel style={yshift=0.5ex},
	xlabel=$\mathcal{R}_{NF}$(\%),
	ylabel=Counts (Total 400),
	ylabel style={yshift=-3.0ex}
	]
\addplot+[ybar interval,mark=no, fill, opacity=0.5] plot coordinates {
(0.74, 0) (0.98, 0) (1.23, 0) (1.47, 3) 
(1.72, 4) (1.96, 8) (2.21, 9) (2.45, 11) 
(2.70, 16) (2.94, 25) (3.19, 24) (3.43, 36) 
(3.68, 27) (3.92, 49) (4.17, 41) (4.41, 33)
(4.66, 23) (4.90, 22) (5.15, 20) (5.39, 12)
(5.64, 15) (5.88, 10) (6.13, 4) (6.37, 7)
(6.62, 1) (6.96, 0) 
};\addlegendentry{Single, Average $\mathcal{R}_{NF}=4.00\%$}
\addplot+[ybar interval,mark=no, fill, opacity=0.5] plot coordinates {
(0.74, 3) (0.98, 4) (1.23, 4) (1.47, 7) 
(1.72, 18) (1.96, 33) (2.21, 50) (2.45, 44) 
(2.70, 51) (2.94, 50) (3.19, 37) (3.43, 32) 
(3.68, 26) (3.92, 15) (4.17, 9) (4.41, 10)
(4.66, 4) (4.90, 3) (5.15, 0) (5.39, 0)
(5.64, 0) (5.88, 0) (6.13, 0) (6.37, 0)
(6.62, 0) (6.96, 0) 
};\addlegendentry{Ensemble, Average $\mathcal{R}_{NF}=2.79\%$}
\end{axis}
\end{tikzpicture}
\caption{$\mathcal{R}_{NF}$ distributions of update \bert{base}$\rightarrow$\bert{base} on MRPC, with new model being single models fine-tuned with different seeds (blue) or ensemble models (red). We train 20 old and 20 new models to calculate 400 pair-wise \nfr.} 
\label{fig:ensemble-dist}
\end{figure}
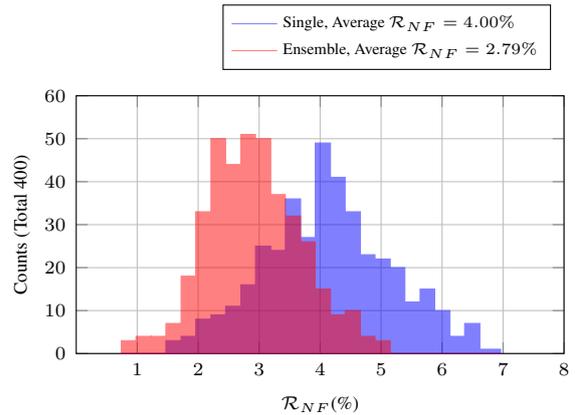

Similar to the findings of \citet{yan2020positive}, we observe in Table \ref{table:CVP} and \ref{table:main-mrpc} that a simple ensemble of models trained with different random initialization before finetuning can reduce regression in some cases. 
We fine-tune \bert{base} on MRPC with $20$ random seeds as our old base models, and another $20$ seeds as our new single models, and another $100$ seeds for building $20$ ensemble models.
Next, we calculate $\mathcal{R}_{NF}$ on the dev set in each model update setup, i.e. $400$ update pairs.
Figure \ref{fig:ensemble-dist} plots their model update regression $\mathcal{R}_{NF}$ distributions.
We observe that the ensemble can not only bring down $\mathcal{R}_{NF}$ but also reduce its variance.

From Figure \ref{fig:ensemble-dist}, we conjecture that each single model could learn a subset of all possible patterns in the data to achieve comparable accuracy on the task.
Models fine-tuned with different seeds could rely on different sets of patterns for predictions, leading to behavioral difference and regression.
On the other hand, ensemble aggregates distinct and complementary behaviors from individual models, leading to less eccentric behavior and increased compatible with individual models on average.
In a parallel work, \citet{allen2020towards} provides a theoretical framework of how ensemble works from the multi-view perspective.
They show that single models can pick up multiple but different views of the data,
and the ensemble naturally collects more view features, leading to a higher accuracy.
Our hypothesis concurs with their findings.

\begin{table}[t!]
\resizebox{0.48\textwidth}{!}{
\begingroup
\renewcommand{\arraystretch}{1.15}
\centering
\begin{tabular}{cc|cc}
\toprule
&&\textbf{Acc}&$\mathcal{R}_{NF}$\\
\midrule
\textbf{Old: \bert{base}} &Single&85.81($\pm$1.07)\% &-\\
\midrule
 &Single& 85.39($\pm$1.43)\% &4.30($\pm$1.53)\%\\
\textbf{$\rightarrow$\bert{base}}&Ensemble&86.18($\pm$1.12)\%&3.08($\pm$1.14)\%\\
&Centric&87.75\%&2.79($\pm$0.69)\%\\
\midrule
&Single&86.32($\pm$2.50)\%&5.37($\pm$2.69)\%\\
\textbf{$\rightarrow$\bert{large}}&Ensemble&	87.65($\pm$1.34)\%&3.64($\pm$1.21)\%\\
&Centric&	87.25\%	&4.24($\pm$0.78)\%\\
\bottomrule
\end{tabular}
\endgroup
}

\caption{The selected single model \textit{centric} can achieve similar accuracy and negative flip as ensemble.}
\label{table:ensemble}
\end{table}

\begin{table*}[]
\small
\resizebox{0.99\textwidth}{!}{
\begingroup
\renewcommand{\arraystretch}{1.15}
\begin{tabular}{l|c|cc|cccc}
\toprule
&\makecell{\textbf{Old: \bert{base}}\\(Error \%)}
&\multicolumn{6}{c}{\textbf{New: \bert{large}} (\textbf{$\mathcal{R}_{NF}$})}\\
\midrule
& 1 Seed & 1 Seed & KD 1 Seed & 5 Seeds & KD 5 Seeds & Centric & Ensemble\\ 
\midrule
\midrule
Coref - He/She  &0.0\%  &13.4\% &22.4\% &12.3($\pm$16.1)\%  &41.2($\pm$47.0)\% &0.0 &0.0($\pm$0.0)\%\\
Vocab - People 
&0.0\%    	& 13.8\%    &\textbf{1.9}\%     &59.5($\pm$34.8)\% &\textbf{34.1}($\pm$47.1)\% &89.0\% &55.2($\pm$50.8)\%\\
Vocab - More/Less
&100.0\%  &0.0\% &0.0\%     &0.0($\pm$0.0)\%  &0.0($\pm$0.0)\%  &0.0\%&0.0($\pm$0.0)\% \\
Taxonomy - Synonym
&0.0\%    	&42.3\%      &\textbf{0.0}\%     &77.0($\pm$26.2)\% &\textbf{20.3}($\pm$44.5)\%  &73.1\% &61.9($\pm$54.1)\%\\
SRL - Pharaphrase 
&0.0\%    &12.5\%        &99.9\%     &63.2($\pm$43.4)\%   &\textbf{42.0}($\pm$48.3)\%  &26.9\%   &61.9($\pm$54.1))\% \\
SRL - Asymmetric Order
&0.0\%  &47.1\%         &\textbf{0.0}\%     &70.7($\pm$20.0)\%  &\textbf{22.3}($\pm$43.7)\%  &92.0\%  &58.6($\pm$52.2)\% \\
SRL - Active/Passive 1 
&0.1\%    &9.3\%          &65.2\%    &58.0($\pm$43.2)\%   &\textbf{52.7}($\pm$50.1)\% &98.5\% &42.4($\pm$51.5)\% \\
SRL - Active/Passive 2
&0.1\% &90.0\%          &\textbf{0.7}\%     &95.5($\pm$4.8)\%  &\textbf{23.8}($\pm$43.3)\%     &99.6\%  &65.2($\pm$56.5)\%   \\
SRL - Active/Passive 3 
&99.9\%  &0.1\%           &0.0\%     &0.1($\pm$0.0)\%    &\textbf{0.0}($\pm$0.0)\%    &0.1\% &0.1($\pm$0.1)\% \\
Temporal - Before/After 
&100.0\% &0.0\%            &0.0\%     &0.0($\pm$0.0)\%   &0.0($\pm$0.0)\%  &0.0\%   &0.0\%   \\
\hline
Average &38.5\%	&19.9\%	&\textbf{14.8}\%	&38.2\%	&\textbf{20.3}\%&43.8\%&33.3\%   \\
\bottomrule
\end{tabular}
\endgroup
}
\caption{Behavioral tests with \textsc{CheckList}. Second column shows the error rate of the old model. 
Remaining columns are \nfr compared with the old model.
The columns with \textit{1 Seed} represent the results with random seed equals to $0$, while columns with \textit{5 seeds} represent 5 seed average.
The columns with \textit{KD} are the models after distillation.
The columns with \textit{Centric} means the single selected with the method mentioned in Section \ref{lab:center}.
}
\label{table:checklist-large}
\end{table*}

\begin{figure}[t!]
\centering
\includegraphics[width=0.46\textwidth]{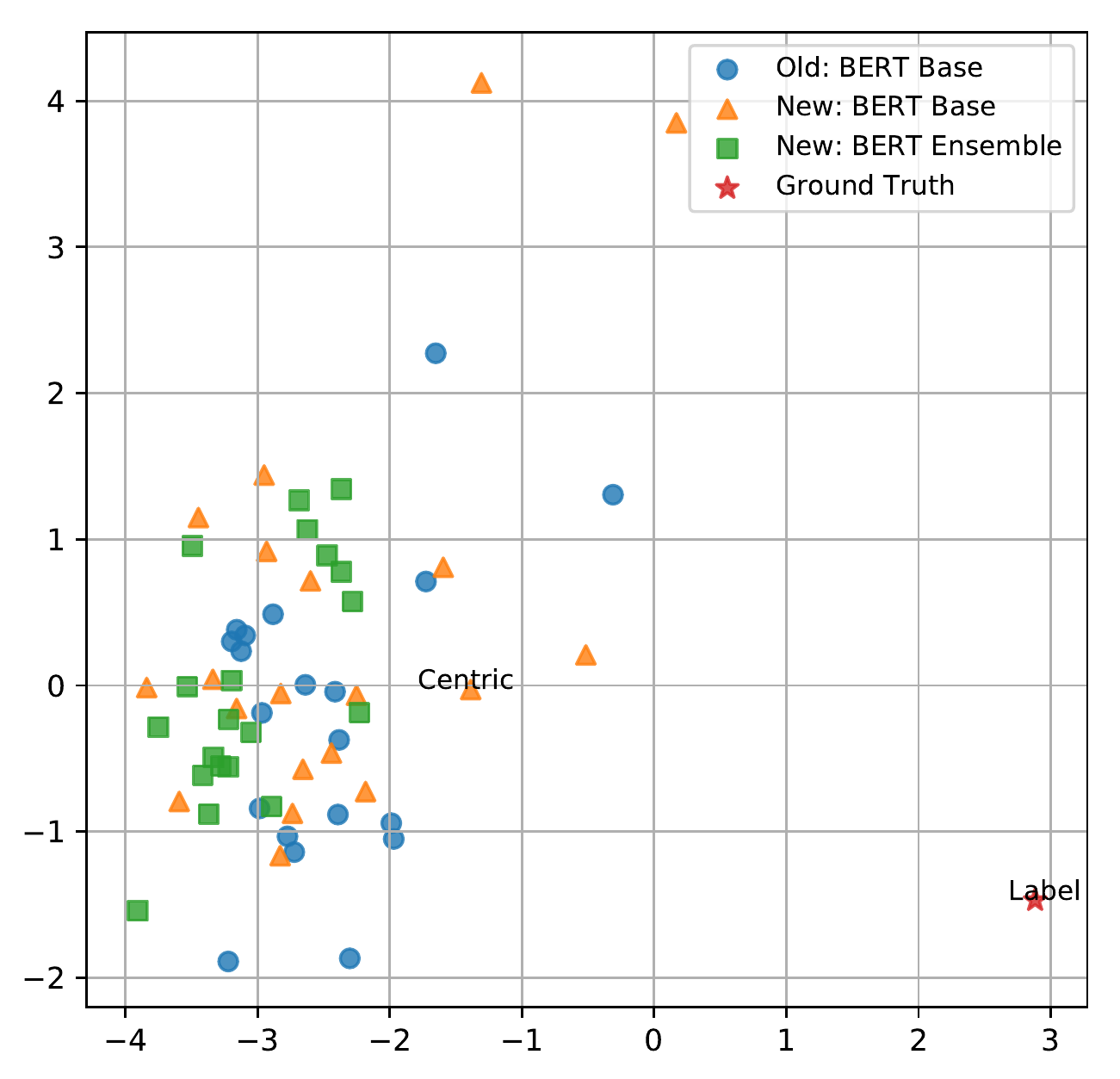}
\caption{2D visualization by PCA of old, new single, and new ensemble models based on dev set predictions.}
\label{fig:ensemble-base}
\end{figure}

However, ensemble is not required to achieve moderate model behavior. To verify this, we designed the following simple model selection procedure.
We first train $20$ new single models, among which we compute for each model the average \nfr on the first half of dev set when updating from the other $19$ models.
We then select the single model with the lowest average \nfr as the \textit{centric}.
Results in Table \ref{table:ensemble} show the accuracy and \nfr on the second half of the dev set. 
Indeed the single \textit{centric} model achieves substantial reduction in \nfr comparable to model ensemble.
We further plot all the \bert{base} models based on their class predictions down-projected by PCA \citep{hotelling1933analysis}.
Figure \ref{fig:ensemble-base} shows that single models tend to spread while ensembles are more concentrated and close together.
We can also see that the \textit{centric} indeed sits near the center of single model cluster.

\subsection{Analyzing Regression with Linguistic Behavioral Testing}

To further understand where the regression happens and how the above methods contribute to reducing regression, we conduct qualitative analysis across diverse linguistic phenomena.
More precisely, we leverage the \textsc{CheckList}  \citep{ribeiro-etal-2020-beyond} behavioral testing and construct regression sets for relevant linguistic capabilities and tests based on perturbations and provided templates.
For example, to test the capability of dealing with lexical taxonomy in the paraphrase detection task, we replace adjectives in one sentence with their synonyms with the label unchanged and expect the model can still predict correctly.
We manually set the templates, apply \textsc{CheckList} to automatically generate testing sentence pairs, and calculate $\mathcal{R}_{NF}$ for each linguistic test.
Detailed linguistic behavioral testing setups with examples can be found in Appendix C.

Table \ref{table:checklist-large} shows the linguistic behavioral testing results when updating from \bert{base} - 1 Seed to \bert{large}.
Each row denotes one specific behavioral test and $500$ cases are sampled in each test.
We focus on negative flips where the new model fails the test while the old model passes.
We can observe that the vanilla fine-tuned \bert{large} has significant regressions on switching with synonyms, asymmetric ordering, and active-passive swap related to people names (see Appendix C).
Also, we observe that models tend to either fail or pass almost all cases in a test, which leads to high variances in $\mathcal{R}_{NF}$.
This implies that models fine-tuned with different seeds can have different behavioral patterns, which could be one source of regression.

Furthermore, Table \ref{table:checklist-large} shows that the distillation can effectively reduce regressions across almost all types of behavioral tests. 
This demonstrates that minimizing the surrogate regression measure, formulated as a knowledge distillation objective, reduces the regression through actually aligning new model's behavior with the old model.

For the ensemble, although it can reduce significant regressions in the benchmark, we observe that it can only improve the model update compatibility on a handful of capabilities.
We hypothesize that the ensemble mostly improves the compatibility with the underlying constituent models.
Without an explicit alignment, it cannot proactively reduce the regression on certain behavior tests when updating from other distinct models.

\section{Related work}

\subsection{Model Update Regression and Solutions}

The backward compatibility representation learning first comes into sight in \citet{Shen2020TowardsBR} on learning inter-operabile visual embeddings for image retrieval tasks.
Later, \citet{yan2020positive} formalize the model update regression problem in machine learning and explore solutions on image classification tasks. 
They suggest negative flip (NF) as the empirical measurement of regression and propose a specialized knowledge distillation loss \citep{Hinton2015DistillingTK} as a surrogate of regression for joint optimizations.
Our work investigates the model update regression in NLP classification tasks, which involve discrete signals and rich linguistic structures. 
We formulate our solutions from the perspective of constraint satisfaction and verify their efficacy on scenarios including update to distinct architectures.

\subsection{Transfer and Lifelong Learning}

Pre-training a model on large corpora and fine-tuning on downstream tasks has emerged as a standard paradigm in NLP \citep{devlin-etal-2019-bert,Lan2020ALBERTAL:,lample2019cross,Raffel2019ExploringTL,brown2020language,Clark2020ELECTRA:}.
Our work follows this transfer learning paradigm but our main focus is to investigate the regression phenomenon when updating backbone pre-trained models.
Another related stream of research is lifelong learning \citep{lopez2017gradient,yoon2017lifelong,de2019continual,sun2019lamol,chuang2020lifelong} or incremental learning \citep{rebuffi2017icarl,chaudhry2018riemannian,prabhu2020gdumb}, which aims to accumulate knowledge learned in previous tasks.
The model update regression problem differs in that models are trained on the same task and dataset, but we update from one model to another.

\subsection{Behavioral Testing of NLP Models}

To analyze whether a fine-tuned model can handle linguistic phenomena for a specific end task, perturbation techniques are often used \citep{belinkov2017synthetic,ribeiro2018semantically,prabhakaran2019perturbation,wu2019errudite,talmor2020olmpics}. 
In particular, \textsc{CheckList} \citep{ribeiro-etal-2020-beyond} leverages and expands those techniques to efficiently evaluate a wide range of linguistic behavioral capabilities of NLP models.
Our work applies \textsc{CheckList} to inspect where the model update regressions come from and on which linguistic phenomena our proposed solutions help to reduce regressions.

\section{Conclusion}

In this work, we investigated the regression in NLP model updates on classification tasks and show that it has a prevalent presence across tasks and models.
We formulated the regression-free model update problem as a constrained optimization problem and reduce it into a joint learning objective on target task while distilling from the old model.
Together with the ensemble, these methods can cut down the regression by $60\%$ at best.
Experiments on the GLUE benchmark showed that ensemble can be effective in reducing the regression when updating to homogeneous models.
On the other hand, knowledge distillation produced more significant regression reductions under the heterogeneous setting.
Through linguistic behavioral testing we showed that distillation can reduce the regression across a wider range of linguistic phenomena than ensemble method. 
While the regression reduction achieved by the discussed methods are promising, they are far from reaching regression-free. 
We leave the design of more advanced regression-reduction methods as future works.

\bibliographystyle{acl_natbib}
\bibliography{BCT,anthology}

\clearpage
\appendix

\section{Full Results of Regression Between SOTA Model Updates}\label{app:regressions-full}

\begin{table*}
\footnotesize
\centering
\resizebox{0.999\textwidth}{!}{
\begingroup
\renewcommand{\arraystretch}{1.25}
\begin{tabular}{cc|cccccccc}
\toprule
 && \textbf{CoLA} & \textbf{SST-2} & \textbf{MRPC} & \textbf{QQP} & \textbf{MNLI-m} & \textbf{MNLI-mm} & \textbf{QNLI} & \textbf{RTE} \\
\midrule
\textbf{Train size}&  & 8.6k & 67k   & 3.7k  & 360k  & 390k  & 390k  & 100k  & 2.5k \\
\textbf{Dev size}& & 1k  & 0.9k  & 0.4k  & 40k   & 9.8k  & 9.8k  & 5.5k  & 0.3k \\
\midrule
\textbf{Old: \bert{base}}& Acc &82.84\%&	92.20\% & 86.03\%&	90.76\%&	83.82\%&	84.13\%&	91.07\%&	67.15\%\\
\midrule
\textbf{$\rightarrow$\bert{base}}& Acc & 83.80\%&	91.93\%& 86.03\%&	90.56\%&	83.55\%&	83.94\%&	90.65\%&	63.18\%\\
& ${\mathcal{R}_{NF}}$& 3.36\%&	2.10\%&	4.17\%&	2.32\%&	3.56\%&	3.67\%&	2.35\%&	11.43\%      \\
\midrule
\textbf{$\rightarrow$\bert{large}}& Acc &85.43\%&	93.23\%&	87.75\%&	91.11\%&	86.10\%&	86.49\%&	92.53\%&	66.43\%\\
& ${\mathcal{R}_{NF}}$        & 3.16\%        & 1.95\%         & 5.88\%        & 2.82\%       & 3.95\%        & 3.69\%        & 2.64\%        & 12.27\%      \\
\midrule
\textbf{$\rightarrow$\textsc{RoBERTa}$_{base}$}& Acc &84.85\%&	94.11\%&	89.22\%&	91.25\%&	87.58\%&	87.74\%&	92.71\%&	63.17\%   \\
& ${\mathcal{R}_{NF}}$        &4.67\%&	1.22\%&	4.66\%&	1.98\%&	2.64\%&	2.38\%&	1.74\%&	13.1\%\\
\midrule
\textbf{$\rightarrow$\textsc{ELECTRA}$_{base}$} &Acc &85.81\%&	95.41\%&	86.03\%&	91.35\%&	88.87\%&	88.67\%&	93.30\%&	72.92\% \\
&${\mathcal{R}_{NF}}$        & 5.18\%        & 1.38\%         & 5.39\%        & 3.20\%       & 3.50\%        & 3.57\%        & 2.65\%        & 7.58\%       \\
\midrule
\textbf{$\rightarrow$\textsc{ALBERTA}$_{base}$} &Acc &76.51\%&	91.86\%&	86.27\%&	90.73\%&	85.26\%&	85.14\%&	91.67\%&	74.73\%\\
&${\mathcal{R}_{NF}}$& 10.74\%&	3.67\%&	6.86\%&	3.78\%&	5.22\%&	5.24\%&	3.70\%&	9.03\%\\
\midrule
\midrule
\textbf{Old: \bert{large}}& Acc &85.43\%&	93.23\%&	87.75\%&	91.11\%&	86.10\%&	86.49\%&	92.53\%&	66.43\%\\
\midrule
\textbf{$\rightarrow$\bert{large-wwm}}&Acc&85.14\%&	94.15\%&	87.01\%&	91.52\%&	86.75\%&	87.24\%&	93.34\%&	70.76\%\\
&${\mathcal{R}_{NF}}$&5.05\%&	1.60\%&	7.82\%&	3.03\%&	5.08\%&	4.67\%&	2.75\%&	15.22\%\\
\midrule
\midrule
\textbf{Old: \textsc{RoBERTa}$_{base}$}& Acc&84.85\%&	94.38\%&	87.25\%&	91.28\%&	88.24\%&	87.63\%&	92.51\%&	70.76\%\\
\midrule
\textbf{$\rightarrow$\textsc{ELECTRA}$_{base}$}&Acc&	85.81\%&	95.41\%&	86.03\%&	91.35\%&	88.87\%&	88.67\%&	93.30\%&	72.92\%\\
&${\mathcal{R}_{NF}}$&
4.31\%&	1.49\%&	6.62\%&	2.77\%&	3.47\%&	3.48\%&	2.62\%&	7.58\%\\
\bottomrule
\end{tabular}
\endgroup
}
\caption{Accuracy and regression measures of different model update variants on GLUE benchmark. \textsc{bert}$_{large-wwm}$ represents the whole-word-masking version.}
\label{table:regression-app}
\end{table*}

Due to the page limitation, we present the full regression update comparison between commonly used pre-trained model pairs \citep{devlin-etal-2019-bert, Liu2019RoBERTaAR, Lan2020ALBERTAL:, Clark2020ELECTRA:} in Table \ref{table:regression-app}.

We show regression in model updates from \bert{base} to the other common used pre-trained models; we also show the regression in updates from \bert{large} to \bert{large-whole-word-masking} and updating from \textsc{RoBERTa}$_{base}$ to \textsc{ELECTRA}$_{base}$.

Other than the universal presence of the regression, Table \ref{table:regression-app} shows that:
\begin{enumerate}
    \item \textbf{The more difference in pre-training method, the higher regression can be observed.} From the results of updating from \bert{base} experiment group we can have this conclusion: The updating factors can increase regression in ascending order: hyperparameters (to \bert{base}), pre-train settings(to \textsc{RoBERTa}$_{base}$), model size (to \bert{large}), pre-train objection (to \textsc{ELECTRA}$_{base}$), model structure(to \textsc{ALBERT}$_{base}$) 
    \item \textbf{Similar updating factor results in similar level of regression.} Updating from \bert{large} to \bert{large-wwm} have similar regression level as in updating from \bert{base} to \textsc{ELECTRA}$_{base}$, both are updating in the pre-training objective. Similarly, Updating from \textsc{RoBERTa}$_{base}$ to \textsc{ELECTRA}$_{base}$ have similar regression as in updating from \bert{base} to \textsc{ELECTRA}$_{base}$.
\end{enumerate}

\section{Selection of Regression Set During Training}
Here, we explore the difference of regressing set selection during knowledge distillation.

For the regression set used in the model training process, we propose several options:
\begin{enumerate}[itemsep=2pt,topsep=2pt,parsep=0pt,partopsep=0pt]
    \item Take the entire training set as our the regression set in training $\mathcal{D}_{reg} = \mathcal{D}_{train}$
    \item Training examples where the old model makes correct predictions $\mathcal{D}_{reg} = \mathcal{D}_{correct}$
    \item Training examples where the old model gets a higher predict probability on the ground-truth class than the new model $\mathcal{D}_{reg} = \mathcal{D}_{better}$, equivalent to adjusting $\alpha$ dynamically according to the performance of the two models, we set $\alpha$ to zero when $p_{\phi_{old}}(y|x) < p_{\phi_{new}}(y|x)$
    \item Extra data from other tasks $\mathcal{D}_{reg} = \mathcal{D}_{extra}$
    \item User-provided regression set, which includes examples with high-stakes $\mathcal{D}_{reg} = \mathcal{D}_{user}$
\end{enumerate}
\noindent We experiment with all options except for the user-provided regression set, see Table \ref{table:mrpc-regression-set}.

Dynamically adapting the regression set according to the current performance of the new model in \textbf{Distillation ($\mathcal{R}_{\text{KL-div}}, \mathcal{D}_{better}$)} offers the most reduction in the regression without sacrificing the accuracy.
We conjecture that it's because we apply the soft regression-free constraint loss precisely on examples where the new model's performance is behind. 
\begin{table*}[t!]
\centering
\begin{tabular}{c|cc|cc}
\toprule
\textbf{Old: \bert{base}}   &\multicolumn{4}{c}{\textbf{Acc}: 86.03\% }   \\\midrule
\textbf{New:} &\multicolumn{2}{c|}{\textbf{$\rightarrow$\bert{base}}} &\multicolumn{2}{c}{\textbf{$\rightarrow$\bert{large}}}\\
& {\textbf{Acc}} &{\textbf{$\mathcal{R}_{NF}$}}  &{\textbf{Acc}} &{\textbf{$\mathcal{R}_{NF}$}}   \\
\midrule

\textbf{Baseline}  & 86.03\%    & 4.17\%     &87.75\% &5.88\% \\
\midrule
\textbf{Distillation($\mathcal{R}_{\text{KL-div}},\mathcal{D}_{train}$)}     & \textbf{87.25\%}    & 2.94\%      &88.73\% &	3.68\%  \\
\textbf{Distillation($\mathcal{R}_{\text{KL-div}}, \mathcal{D}_{correct}$)}& 87.01\%    & 3.92\%&88.48\% &4.90\%\\
\textbf{Distillation($\mathcal{R}_{\text{KL-div}}, \mathcal{D}_{better}$)}     & 87.01\%    & \textbf{1.72\%}&88.73\% &2.45\%\\
\textbf{Distillation($\mathcal{R}_{l_2}, \mathcal{D}_{better}$)}    &85.29\% &2.45\% &88.73\%    &\textbf{2.21\%}\\
\bottomrule
\end{tabular}
\caption{Results on MRPC of our proposed techniques towards regression-free model updates. The model is updated from\bert{base} to \bert{base}, e.g. change fine-tune setups, or \bert{large}. 
$\mathcal{D}_{(\cdot)}$ denotes the regression set for joint training with distillation and classification. }
\label{table:mrpc-regression-set}
\end{table*}

\section{Linguistic Behaviour Test settings}\label{app:checklist}

In the linguistic behaviour tests, we go through a variety range of linguistic aspects and design test examples following CheckList\citep{ribeiro-etal-2020-beyond}.

In Table \ref{table:checklist-detail} we show the tests for linguistic behaviour tests. 
Please find the example test cases in the third column for each testing.

\begin{table*}[t!]
\resizebox{0.98\textwidth}{!}{
\begingroup
\renewcommand{\arraystretch}{1.15}
\begin{tabular}{l|p{6cm}|p{8cm}l}
\toprule
\textbf{Category}&\textbf{Description}&\textbf{Example}& \textbf{Label}\\
\midrule
Coref - He/She    &Reverse he or she. 
& If Charles and Jessica were alone , do you think \textbf{he} would reject \textbf{her}?    &False\\
&&If Charles and Jessica were alone , do you think \textbf{she} would reject \textbf{him}?\\\midrule
Vocab - People    &Add modifiers that preserve sentence
& Wendy is friendly to Kevin.&True\\
&semantics.& Wendy is \textbf{truely} friendly to Kevin.\\\midrule
Vocab - More/Less      & Swap more with less.  
&I can become \textbf{more} passive. &True\\
&&I can become \textbf{less} passive. \\\midrule
Taxonomy - Synonym     &Replace synonym.
&I can become more \textbf{courageous}.    &True  \\
&&I can become more \textbf{brave}.\\\midrule

SRL - Pharaphrase &  Somebody think $\rightarrow$ According to Somebody.
&Who do conservatives think is the happiest surgeon in the world ?   &True\\
&& Who is the happiest surgeon in the world according to conservatives ? \\\midrule
SRL - Asymmetric Order      &Order   does matter for asymmetric  
& \textbf{Shannon} is proposing to \textbf{Samantha}.&False\\
&relations.& \textbf{Samantha} is proposing to \textbf{Shannon}. \\\midrule
SRL - Active/Passive 1    &Traditional   SRL: active / passive swap
&Jeremy missed the game. &True\\
&&The game was missed by Jeremy.\\\midrule
SRL - Active/Passive 2   &Traditional   SRL: active / passive swap.
&  Christian remembers Alyssa. &True \\
&with people.&Alyssa is remembered by Christian.\\\midrule
SRL - Active/Passive 3    &Traditional   SRL: wrong active / passive   
&Sara \textbf{took} the castle.&False\\
&swap.&Sara \textbf{was taken} by the castle. \\\midrule


Temporal - Before/After & Before becoming somebody $\rightarrow$ after becoming somebody.
&What was Noah Myers 's life \textbf{before} becoming an architect ? &False \\
&&What was Noah Myers 's life \textbf{after} becoming an architect ?\\\midrule


\bottomrule
\end{tabular}
\endgroup
}
\caption{Regression Tests details with CheckList.}
\label{table:checklist-detail}
\end{table*}

\section{GLUE Details}
\label{apx:glue}

The GLUE datasets are described as follows\cite{Jiao2020TinyBERTDB}:

\noindent\textbf{MNLI.} Multi-Genre Natural Language Inference is a large-scale, crowd-sourced entailment classification task \citep{williams-etal-2018-broad}. Given a pair of $\langle premise, hypothesis \rangle$, the goal is to predict whether the $hypothesis$ is an entailment, contradiction, or neutral with respect to the $premise$.

\noindent\textbf{QQP.} Quora Question Pairs is a collection of question pairs from the website Quora. The task is to determine whether two questions are semantically equivalent \citep{chen2018quora}.

\noindent\textbf{QNLI.} Question Natural Language Inference is a version of the Stanford Question Answering Dataset which has been converted to a binary sentence pair classification task by \citet{wang2018glue}. Given a pair $\langle question, context \rangle$. The task is to determine whether the $context$ contains the answer to the $question$.

\noindent\textbf{SST-2.} The Stanford Sentiment Treebank is a binary single-sentence classification task, where the goal is to predict the sentiment of movie reviews~\cite{socher2013recursive}.

\noindent\textbf{CoLA.} The Corpus of Linguistic Acceptability is a task to predict whether an English sentence is a grammatically correct one~\citep{warstadt2019neural}.

\noindent\textbf{STS-B.} The Semantic Textual Similarity Benchmark is a collection of sentence pairs drawn from news headlines and many other domains \citep{cer2017semeval}. The task aims to evaluate how similar two pieces of texts are by a score from 1 to 5.

\noindent\textbf{MRPC.} Microsoft Research Paraphrase Corpus is a paraphrase identification dataset where systems aim to identify if two sentences are paraphrases of each other~\cite{dolan2005automatically}.

\noindent\textbf{RTE.} Recognizing Textual Entailment is a binary entailment task with a small training dataset~\citep{bentivogli2009fifth}.

\end{document}